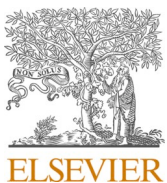
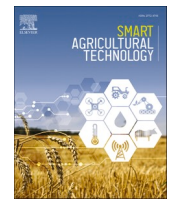

# Mob-based cattle weight gain forecasting using ML models

Muhammad Riaz Hasib Hossain [a,*], Md Rafiqul Islam [b], S.R. McGrath [c], Md Zahidul Islam [d], David Lamb [e,f]

[a] *School of Computing, Mathematics and Engineering, Charles Sturt University, Wagga Wagga, NSW 2650, Australia*
[b] *School of Computing, Mathematics and Engineering, Charles Sturt University, Albury, NSW 2640, Australia*
[c] *Gulbali Institute for Agriculture, Water and Environment, Charles Sturt University, Wagga Wagga, NSW 2678, Australia*
[d] *School of Computing, Mathematics and Engineering, Charles Sturt University, Bathurst, NSW 2795, Australia*
[e] *Precision Agriculture Research Group, University of New England, NSW 2351, Australia*
[f] *Food Agility CRC Ltd, Sydney, NSW 2000, Australia*

ABSTRACT

Forecasting mob-based cattle weight gain (MB-CWG) may benefit large livestock farms, allowing farmers to refine their feeding strategies, make educated breeding choices, and reduce risks linked to climate variability and market fluctuations. In this paper, a novel technique termed MB-CWG is proposed to forecast the one-month advanced weight gain of herd-based cattle using historical data collected from the Charles Sturt University Farm.[1] This research employs a Random Forest (RF) model, comparing its performance against Support Vector Regression (SVR) and Long Short-Term Memory (LSTM) models for monthly weight gain prediction. Four datasets were used to evaluate the model's performance, using 756 sample data from 108 herd-based cattle, along with weather data (rainfall and temperature) influencing CWG. The RF model performs better than the SVR and LSTM models across all datasets, achieving an $R^2$ of 0.973, RMSE of 0.040, and MAE of 0.033 when both weather and age factors were included. The results indicate that including both weather and age factors significantly improves the accuracy of weight gain predictions, with the RF model outperforming the SVR and LSTM models in all scenarios. These findings demonstrate the potential of RF as a robust tool for forecasting cattle weight gain in variable conditions, highlighting the influence of age and climatic factors on herd-based weight trends. This study has also developed an innovative automated pre-processing tool to generate a benchmark dataset for MB-CWG predictive models. The tool is publicly available on GitHub[2] and can assist in preparing datasets for current and future analytical research.

## 1. Introduction

Cattle production contributes significantly to both global food security and agricultural economies, requiring efficient livestock management practices for optimal productivity and sustainability. As the world's population grows, there is a rising demand for high-quality protein sources, placing significant pressure on the livestock industry. Projections indicate that livestock demand will rise to $20 billion by 2026, supported by a stable yearly increase of approximately 0.3 % observed since 2017 [1]. Meeting this demand requires innovative strategies to ensure sustainable livestock production while maintaining profitability. Effective management of resources, including predictive tools for livestock growth, has become critical to addressing these challenges [2]. Accurate forecasting of cattle weight gain could assist farmers in optimising management strategies and marketing plans to achieve greater profitability.

MB-CWG describes the increase in the weight of a herd of cattle during a specific timeframe, serving as a crucial metric for farms to assess the progress and maturity of cattle [3–5]. MB-CWG could be an important tool in modern livestock production systems for several reasons. Firstly, it facilitates precision feeding strategies, allowing farmers to optimise nutrition based on predicted weight gain patterns, leading to improved cattle health and cost-effective resource utilisation [6]. Additionally, accurate weight gain predictions empower farmers to make informed decisions regarding breeding programs. Moreover, forage conditions are significantly influenced by climate-related factors such as variability in rainfall and temperature, and these changes have a direct effect on the weight gain trajectory of grazing animals [7]. Market conditions such as cattle prices, feed costs, and consumer demand can lead farmers to modify key management decisions. These adjustments, including timing of sales, investment in supplementary feeding, and culling decisions, can all influence weight gain patterns [8]. Ultimately, MB-CWG prediction could be instrumental in enhancing overall farm productivity, sustainability, and the economic viability of cattle farming operations.

Despite significant advancements in ML applications for agriculture, limited research has been conducted on MB-CWG forecasting. Existing studies often focus on individual animal predictions, which may not reflect the aggregated dynamics of mob-based systems [9,10]. This gap highlights the need for robust, scalable models that can handle group-level data while maintaining accuracy. This study addresses this gap by applying and comparing three leading ML models: RF, SVR, and LSTM, on datasets that include different combinations of cattle background information, weather factors, and age-related data. The findings serve as a practical resource for livestock managers and researchers to manage cattle and improve overall productivity in mob-based grazing systems. This paper has made the following key contributions.

- This study integrated weather and historical data with real-time weight measurements from the Optiweigh system at CSU Digital Farm, aiming to reduce stress on cattle during the data collection phase.
- A novel tool has been developed to transform time-series cattle data into non-time-series experimental data, making it suitable for ML models within the MB-CWG forecasting framework.
- A comparative analysis of ML models (including a DL model) has been performed in this experiment to find a more effective model for predicting MB-CWG on a monthly basis.

This paper is structured as follows: 1) Section 2 reviews previous studies that have investigated for MB-CWG prediction; 2) Section 3 presents the methodology, including data collection, pre-processing, and the architecture of ML models; 3) Section 4 describes the results of the empirical evaluation; 4) Section 5 discusses the outcome of the research; and 5) the concluding section summarises the study's results and future research direction.

A complete list of abbreviations employed in this research is compiled in Table 1 to facilitate readability.

## 2. Related works

Several researchers have utilised ML and DL models for indirect measurements of cattle weight, leveraging auxiliary data such as morphological features, body dimensions, or spectral imaging rather than directly forecasting MB-CWG. Traditional models like Linear Regression (LR) have been widely used for their simplicity and effectiveness with numerical datasets, as evidenced by Grzesiak et al. [11] and Heinrichs et al. [12]. However, advanced models like RF and Light Gradient Boosting Machine (LightGBM) have emerged to handle increasing data complexity and size. For example, Dang et al. [13] applied LightGBM to a dataset of 33,546 samples, achieving an RMSE of 24.75, while Grzesiak et al. [11], Liu et al. [14], and Na et al. [15] demonstrated the efficacy of RF models. Image-based methods, including those employing CNNs, have revolutionised cattle weight estimation by extracting high-dimensional features from photographs or videos, with studies by Ruchay et al. [16] and Gjergji et al. [17] reporting accuracies exceeding 90 %. Artificial Neural Networks (ANNs) have shown significant promise in numerical data processing, achieving highly accurate results, such as a MAPE of 1.37 % reported by Biase et al. [18]. The Prophet model presents a viable option for addressing non-linear time series trends in MB-CWG forecasting [19]. These developments illustrate the versatility and robustness of ML and DL frameworks in integrating diverse data modalities to advance indirect cattle weight estimation.

Despite several studies on CWG estimation using ML or DL techniques, significant gaps remain in forecasting MB-CWG. MB-CWG forecasting involves unique hurdles, such as synthesising mob-level data, integrating herd movement patterns, and managing the influence of weather conditions and intra-herd variations. Overcoming these challenges necessitates designing novel algorithms and methodologies tailored to the intricacies of mob-based analysis, a domain yet to be fully explored in the existing literature.

Weight gain and its prediction depend on numerous factors and conditions, as highlighted by Nkrumah et al. [20]. Studies have identified key determinants, including feed intake [21,22], genetics [23,24], nutrition [23–25], health [23,24], age [26–28], and gender [26–28]. Among these, feed intake is significant, with its integration into genomic data enhancing growth trait predictions [21,22]. In grazing systems, however, farmers have limited control over feed intake, which depends heavily on pasture quality and weather conditions. Agung et al. [26] demonstrated that cattle of the same breed can exhibit varying rates of weight gain when subjected to different climates, underscoring the critical role of environmental factors such as temperature and humidity. Similarly, Huuskonen & Huhtanen [29] highlighted that weather modulates metabolic energy intake, a primary predictor of weight gain. Despite the findings, there is a significant gap in research focused on the impact of weather on mean MB-CWG. This gap highlights the necessity for further investigation to understand how specific weather variables contribute to variations in MB-CWG, which could inform more effective cattle management practices.

Identifying and filling the inconsistencies in previous research can create influential research. One notable gap in existing studies lies in the emphasis on individual cattle rather than MB-CWG for weight estimation, leaving the domain of future weight forecasting underexplored. While the focus has been mainly on immediate weight assessments, methods for forecasting weight gain in mob settings remain scarce. Variables such as age, breed, feeding practices, and weather are recognised as critical to weight gain. However, these factors are often used indirectly rather than being incorporated into predictive models. Moreover, research typically isolates the effects of seasonal changes and age-related metabolic shifts, neglecting their combined interactions. This study aims to analyse the effectiveness of different models in forecasting weight gain in mob-based cattle systems, relying on sequential live weight measurements and integrating parameters such as cattle age and environmental weather conditions.

**Table 1**
A list of the abbreviations.

| Acronym | Meaning | Acronym | Meaning |
|---|---|---|---|
| ANN | Artificial Neural Networks | MAE | Mean Absolute Error |
| BOM | Bureau of Meteorology | MAPE | Mean Absolute Percentage Error |
| CNN | Convolutional Neural Network | $R^2$ | Coefficient of Determination |
| CV | Cross-Validation | RFID | Radio-Frequency Identification |
| GDF | Global Digital Farm | RF | Random Forest |
| EID | Electronic Identification | RMSE | Root Mean Square Error |
| LightGBM | Light Gradient Boosting Machine | SVR | Support Vector Regression |
| LR | Linear Regression | SOO | Step-On-Off |
| LSTM | Long Short-Term Memory | WoW | Walk-Over Weighing |
| MB-CWG | Mob-Based Cattle Weight Gain | | |

## 3. Methodology

### 3.1. Database samples

#### 3.1.1. Fields of study

This study used a mob of Angus cattle ($n = 267$) from Global Digital Farm (GDF[3]) located at Charles Sturt University in Wagga Wagga, Australia. Data were collected from February 2022 (when cattle were weaned) to October 2022 (when cattle were sold). The birth dates of the cattle ranged from 21st July 2021 to 6th September 2021. Data collection was approved by the Charles Sturt University Animal Ethics Committee.

#### 3.1.2. Data collection

In this study, data were collected from three sources to analyse CWG patterns. The first raw data were derived from the university's database, containing detailed background histories of each animal. The second raw data comprised daily weather information obtained from the Bureau of Meteorology (BOM[4]) for the period from December 2021 to December 2022. The inclusion of two months before and after the primary research window (February to October) was intended to capture delayed weather influences on weight gain dynamics. The third raw data consisted of daily individual cattle weight measurements recorded using the Optiweigh system [30]. Table 2 describes all features collected from these sources. Temperature and rainfall were included as seasonal inputs based on their established influence on cattle growth across different months. Periods of intense heat, especially during summer, are likely to reduce cattle growth through increased thermal stress [31]. Pasture availability, another seasonally driven variable, was not recorded directly but was inferred through weather trends. The dependence of pasture dynamics on rainfall and temperature supported its inclusion through environmental proxies. Subsequently, all collected raw data were integrated into a single dataset, excluding the breed and gender features, as all cattle were Angus-castrated males. The Electronic Identification (EID) was retained solely to identify each animal within the database uniquely.

The Optiweigh device was employed to measure the actual weight of individual cattle. This tool combines a portable weighing platform with a radio-frequency identification (RFID) reader for animal identification and a solar-powered satellite communication unit for real-time data transmission. Fig. 1 illustrates the Optiweigh system process flow. This process involved a portable weighing system designed for cattle that they can freely access in the paddock, using salt or molasses as a motivating factor to encourage cattle to participate voluntarily in the weighing procedure [32]. As cattle step onto the platform with their front legs to reach the feed, the front-end weight is recorded as a reliable estimate of their total weight [33]. The collected data, linked to individual animals via EID, is transmitted in real-time to a cloud-based dashboard. This setup allows for frequent, stress-free monitoring of cattle weight changes.

#### 3.1.3. Data preparation

The data preparation process for MB-CWG forecasting involved cleansing, synthesising, transforming, and structuring the collected raw data [34]. This step ensured data integrity and relevance for the experimental analysis. The preparation process addressed inconsistencies in the raw data, integrated diverse datasets, and structured the information for accurate forecasting.

- Animal weight data processing: A key challenge during data preparation was the inconsistency in daily weight measurements, as not all animals accessed the Optiweigh unit daily. From February 2022 to

---

[3] The paper of GDF ( https://www.csu.edu.au/global-digital-farm/home )
[4] The website of BOM (http://bom.gov.au)

**Table 2**
Summary of collected features.

| SN | Feature | Example | Reasons to Include or Exclude for Training Models |
|---|---|---|---|
| | Data Source: University's database | | |
| 01. | EID | 982123768703781 | Exclude - Acts as an identifier; not directly relevant for predictive modelling. |
| 02. | Breed | Angus | Exclude - Include only a single breed (Angus), offering no variability for the model. |
| 03. | Gender | Male | Exclude - Includes only male cattle, making the feature irrelevant for modelling. |
| 04. | Date of Birth | 8th August 2021 | Include - Required to calculate age, a key factor influencing growth dynamics and model predictions. |
| 05. | Weaning Date | 31st January 2022 | Include - Provides context for growth phases. |
| 06. | Weaning Weight | 194.5 Kg | Include - Serves as a baseline for forecasting weight gain. |
| | Data Source: BOM | | |
| 01. | Rainfall Date | 25th February 2022 | Include - Required to calculate rainfall each day, a key factor influencing growth dynamics and model predictions. |
| 02. | Rainfall Quantity | 16.6 mm | Include - Correlates rainfall patterns with cattle growth that are relevant for growth rate analysis models. |
| 03. | Temperature Date | 18th February 2022 | Include - Required to calculate temperature each day, a key factor influencing growth dynamics and model predictions. |
| 04. | Temperature | 28.7 °C | Include - Correlates temperature patterns with cattle growth that are relevant for growth rate analysis models. |
| | Data Source: Optiweigh | | |
| 01. | Date of Weight | 18th February 2022 | Include - Links weight measurements to precise time frames, critical for tracking growth patterns. |
| 02. | Actual Weight | 209 Kg | Include - Serves as the primary target variable for training ML models. |

October 2022, 118 out of the 267 cattle accessed the unit at least once each month. However, data from 10 cattle were removed due to irregular patterns that deviated significantly from expected biological growth trends. To address noise, outlier detection was performed using a statistical threshold-based approach, which eliminated extreme values likely caused by measurement errors. The remaining 108 cattle were selected for the monthly MB-CWG forecasting. For each of these animals, monthly mean weights were calculated, resulting in a dataset containing nine monthly means for each of the 108 cattle, totalling 972 records (108 cattle × 9 months). As part of transitioning the dataset from time-series to non-time-series format, the first and last records of each cattle were removed. This exclusion was necessary due to the unavailability of weight data from the previous month for February and the following month for October in the raw data. These data points are essential for generating the features necessary to forecast MB-CWG. This step reduced the dataset to 756 records. Subsequently, detailed background information for each animal was integrated into the dataset to enhance its contextual relevance for analysis.

- Weather data integration: Weather data, comprising daily temperature and rainfall, was collected from December 2021 to December

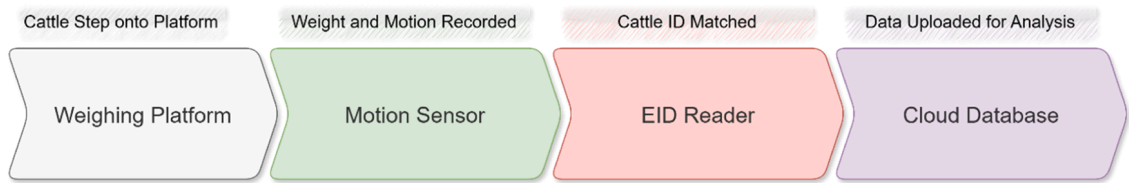

**Fig. 1.** Optiweigh system process flow.

2022. Daily weather observations were aggregated into monthly averages to align with the structure of the animal weight data. The transformation produced a standardised dataset that was integrated with the cattle weight records, creating a robust foundation for subsequent forecasting analyses.

Fig. 2 presents the complete sequence of data preparation activities, from raw data cleaning through integration and feature engineering. Despite the growing popularity of data augmentation for improving model robustness in small datasets [35], it was deliberately not employed in this study. The finalised dataset comprised 756 records, but synthetic expansion was avoided due to concerns that artificially generated data might replicate existing patterns too closely or violate physiological and domain-specific constraints, leading to inaccurate or misleading outcomes. Previous research has highlighted that while augmentation can be effective in image or unstructured text domains, its application to numeric tabular datasets demands careful consideration [36]. Similarly, transfer learning was not applied due to the absence of a sufficiently large dataset. According to Singh et al. [37], transfer learning is typically more effective when pre-trained models originate from similar domains with substantial amounts of labelled data, conditions not met in this research.

Feature extraction, on the other hand, involves creating or transforming existing data to generate informative features that are more meaningful or predictive for the model [38]. In regression, feature extraction is often used to help the model capture important patterns that might not be directly present in the original data [39]. In this study, multiple features were engineered from the existing dataset to train the models, as described below.

- Current month and previous month weight: Current month and previous month weight provide recent weight information, essential

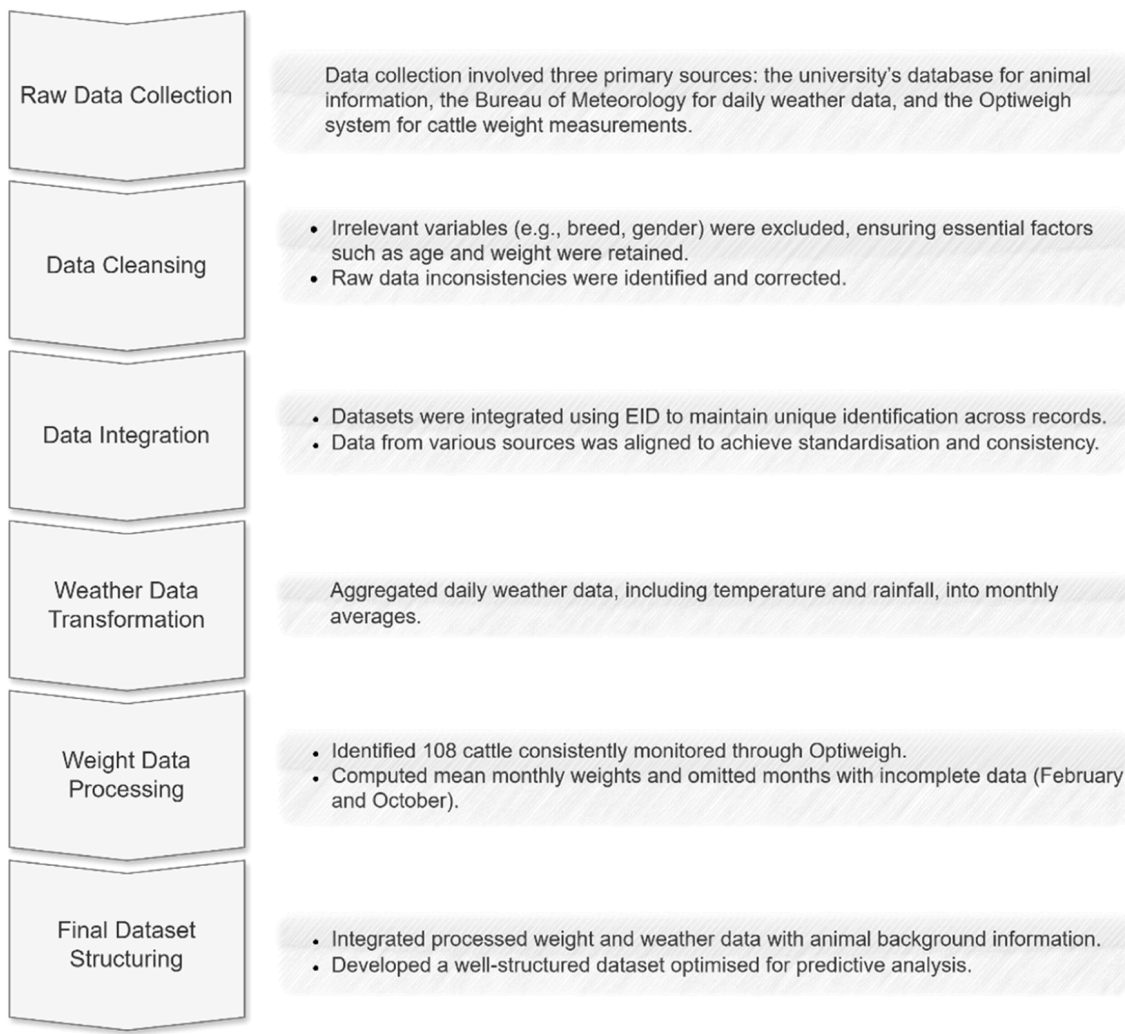

**Fig. 2.** Sequential steps in data preparation process.

for understanding short-term growth trends. Missing values within these records were addressed through a structured pre-processing pipeline using mean-based imputation, ensuring consistency across monthly datasets and maintaining coherence for model training. Although interpolation techniques such as linear or spline interpolation are widely applied in time-series contexts [40], the conversion of data into a non-time-series format in this study made mean imputation the more suitable choice. Other studies have also demonstrated the effectiveness of imputation methods, with multiple imputations generating statistically valid replacements that account for uncertainty in missing data [41].

- Current age by month: Current age by month helps the model adjust predictions for varying growth rates based on cattle maturity. Weight gain patterns differ significantly across age groups, so this feature is critical to achieving accurate forecasts.
- Rainfall (current, one month prior, and two months prior) and temperature (current, one month prior, and two months prior) are included for their significant contributions to seasonal variations in weight gain. During the rainy season, improved pasture quality and reduced thermal stress foster growth, while dry seasons often lead to diminished feed availability and increased temperature extremes, negatively affecting weight gain.

Four distinct datasets were generated from the pre-processed data to analyse varying factors. Table 3 provides a clear comparison of each dataset's feature composition, highlighting how each is tailored for a specific analytical focus. Dataset 01 integrates both weather and age-related variables, offering a comprehensive perspective with features such as weaning weight, current age by month, and rainfall over various time intervals. Dataset 02 includes only weather-related factors, omitting age as a feature, while Dataset 03 reverses this by including age but excluding weather factors. Finally, Dataset 04 provides a baseline by excluding both age and weather factors.

### 3.1.4. Cross-validation (CV) strategy

CV is a fundamental methodology in ML or DL to ensure reliable model evaluation and performance estimation. It operates by iteratively utilising subsets of the dataset for training and validation purposes. This iterative process is crucial for reducing overfitting, identifying optimal hyperparameter values, and maintaining consistent model performance across various data subsets.

This research implemented 10-fold cross-validation to assess the performance of the RF, SVR and LSTM models. This approach assesses accuracy by splitting the dataset into ten equal sections, known as folds. During each iteration, the model is trained using nine of these folds, with the last fold reserved for testing [42,43]. This cycle is repeated ten times, ensuring each fold is used for testing exactly once. By employing this strategy, the research achieved a balanced and thorough evaluation of the ML and DL models' predictive performance.

### 3.2. Statistical analysis

A statistical analysis was performed in this study to identify the primary factors affecting cattle weight, a critical aspect of efficient livestock management. The dataset comprised variables representing growth metrics, weather conditions and age, enabling a comprehensive analysis of potential determinants.

The relationship between age (measured in months) and weight was examined to understand the growth patterns of cattle. Fig. 3 presents a scatterplot showing an evident upward trend, confirming that age strongly influences weight. This finding was supported by calculating the Pearson correlation coefficient, which demonstrated a strong positive association ($p < 0.05$). Further analysis using polynomial regression was explored to capture potential nonlinearities in growth patterns.

The study also investigated how environmental factors such as rainfall and temperature influence cattle weight. The results suggested that higher rainfall and moderate temperatures were associated with greater cattle weight. Multiple regression analysis confirmed these trends, with both rainfall and temperature showing significant predictive power for next month's weight ($p < 0.05$).

### 3.3. Automated tool

An automated data pre-processing system addresses the critical challenge of data variability, ensuring consistent and scalable MB-CWG forecasting. This research introduces an automated data pre-processing system that effectively mitigates issues arising from inconsistent raw data formats and the lack of standardised datasets. By improving the reliability of predictive models, this system significantly enhances the potential for practical applications of MB-CWG research.

CWG forecasting remains a complex task due to variations in nutrition, breed, age, environmental conditions, and inconsistencies across datasets. For example, as highlighted by Sarini et al. [44], age-specific variability contributed significantly to discrepancies in CWG results. Furthermore, the authors Weber et al. [24] indicated that ML methods trained on image-based data from genetically similar cattle failed to achieve consistent outcomes when used on cattle with distinct genetic traits. These insights underline the urgent need for pre-processing systems that can align datasets into a cohesive format. Additionally, most researchers rely on privately collected data due to the limited availability of public datasets. The absence of a benchmark dataset compounds the issue, as models often require fine-tuning for each specific dataset.

This research involved the development of an automated data pre-processing system on the Visual Studio 2022 platform, where C# was used for backend operations and HTML, CSS, and JavaScript were utilised to design its front-end framework. Its modular architecture makes it adaptable to varied research demands while emphasising ease of use for researchers. The software has been shared on GitHub, allowing researchers to download it. The system is expected to contribute significantly to the development of a benchmark dataset, which will be instrumental in improving the generalisability of predictive models.

**Table 3**
Features for each dataset.

| Feature No | Dataset 01: Including Weather and Age Factors | Dataset 02: With Weather Factors (Excluding Age Factor) | Dataset 03: With Age Factor (Excluding Weather Factors) | Dataset 04: Excluding Age and Weather Factors |
|---|---|---|---|---|
| 01 | Weaning weight | Weaning weight | Weaning weight | Weaning weight |
| 02 | Current age by month | Current month weight | Current age by month | Current month weight |
| 03 | Current month weight | Next month weight | Current month weight | Next month weight |
| 04 | Next month weight | Previous month weight | Next month weight | Previous month weight |
| 05 | Previous month weight | Current month rainfall | Previous month weight | |
| 06 | Current month rainfall | Previous first-month rainfall | | |
| 07 | Previous first-month rainfall | Previous second-month rainfall | | |
| 08 | Previous second-month rainfall | Current month temperature | | |
| 09 | Current month temperature | Previous first-month temperature | | |
| 10 | Previous first-month temperature | Previous second-month temperature | | |
| 11 | Previous second-month temperature | | | |

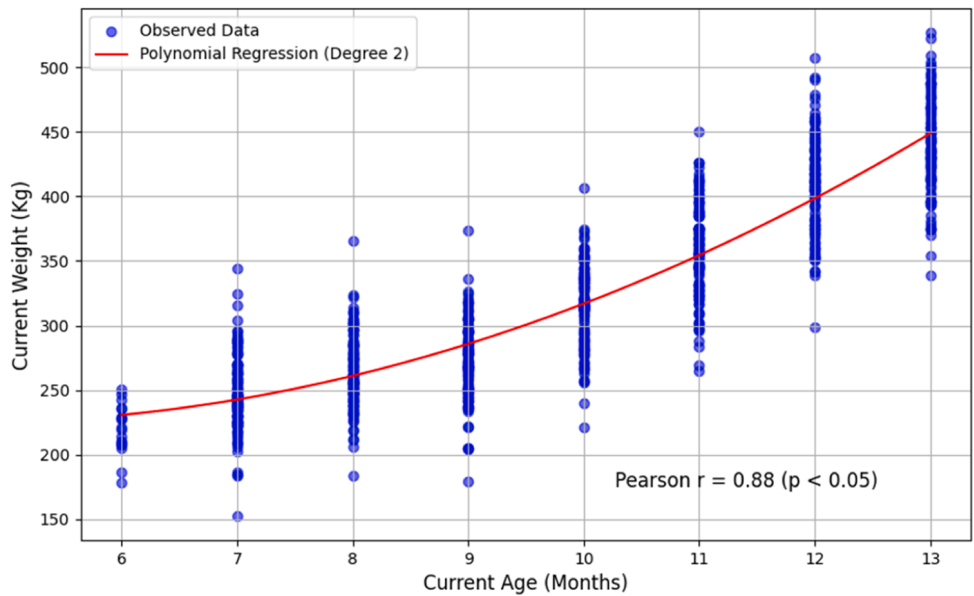

**Fig. 3.** Relationship between age and cattle weight.

The automated data pre-processing tool offers a simplified approach for preparing a dataset for CWG research, enhancing data consistency and reducing the time spent on manual processing. The process begins with user authentication and authorisation through a login system, providing secure access to the tool's features. Following a successful login, an entry form interface (Fig. 4) is displayed, allowing the upload of multiple CSV files containing unprocessed raw data. Users then synchronise the columns of the uploaded files with the predefined structure of the proposed format. After initiating the process by clicking the run button, the application transitions to a final interface (Fig. 5), and the cleaned, formatted dataset is generated. The resulting CSV file is compatible with a variety of ML platforms and includes additional attributes extracted during feature extraction. This functionality enables researchers to focus on model development while ensuring the reliability of CWG datasets.

**Fig. 4.** Entry form of the proposed software.

**Fig. 5.** Result generating form of proposed software.

*3.4. RF, LSTM and SVR models*

The study utilised RF, LSTM, and SVR models to predict MB-CWG for the upcoming month, applying historical and weather data as independent variables. Python-based implementation in Jupyter Notebook utilised predictors such as Birth Weight, Rainfall Date, Weather Temperature, and others to determine "next-month weight". RF is highly effective at capturing non-linear relationships, reducing overfitting, and providing insights into feature importance. SVR offers strong performance in high-dimensional spaces and leverages non-linear kernels to model complex interactions. LSTM retains memory and detects patterns, excelling in modelling developmental trends, even without strict time dependence.

The RF mode extends the Bagging technique by building multiple decision trees in parallel on randomly sampled subsets of the training data, then averaging their outputs to enhance prediction accuracy and robustness [45]. Each tree is constructed independently, reducing overfitting and improving generalisation. This ensemble approach captures diverse data patterns, leading to more stable and reliable forecasts, as illustrated in Fig. 6(a).

LSTM networks specialise in processing sequential data through recurrent links that support long-term information retention [46,47]. Each unit in the network contains input, forget, and output gates, which use sigmoid activation functions to control how data is stored, discarded, or passed along. The cell state, represented in Fig. 6(b) as a horizontal line, carries important information forward, allowing the network to recognise patterns across time-separated inputs.

SVR is rooted in statistical learning theory and aims to optimise a mathematical function, seeking a hyperplane in a high-dimensional space to fit the data best [48]. The primary goal in SVR is to find a regression line or hyperplane that minimises prediction error while maximising the margin around it. Instead of separating data points into distinct classes, SVR tries to keep as many data points as possible within a specified threshold, known as the epsilon tube, around the regression line. Key components of SVR include the kernel function, which allows for mapping input data into higher-dimensional spaces, and the support vectors, or data points that lie outside the epsilon tube, contributing directly to the definition of the regression line. By utilising various kernel functions, such as linear, polynomial, or radial basis functions, SVR is effective in capturing non-linear relationships within data, providing flexibility for different regression scenarios.

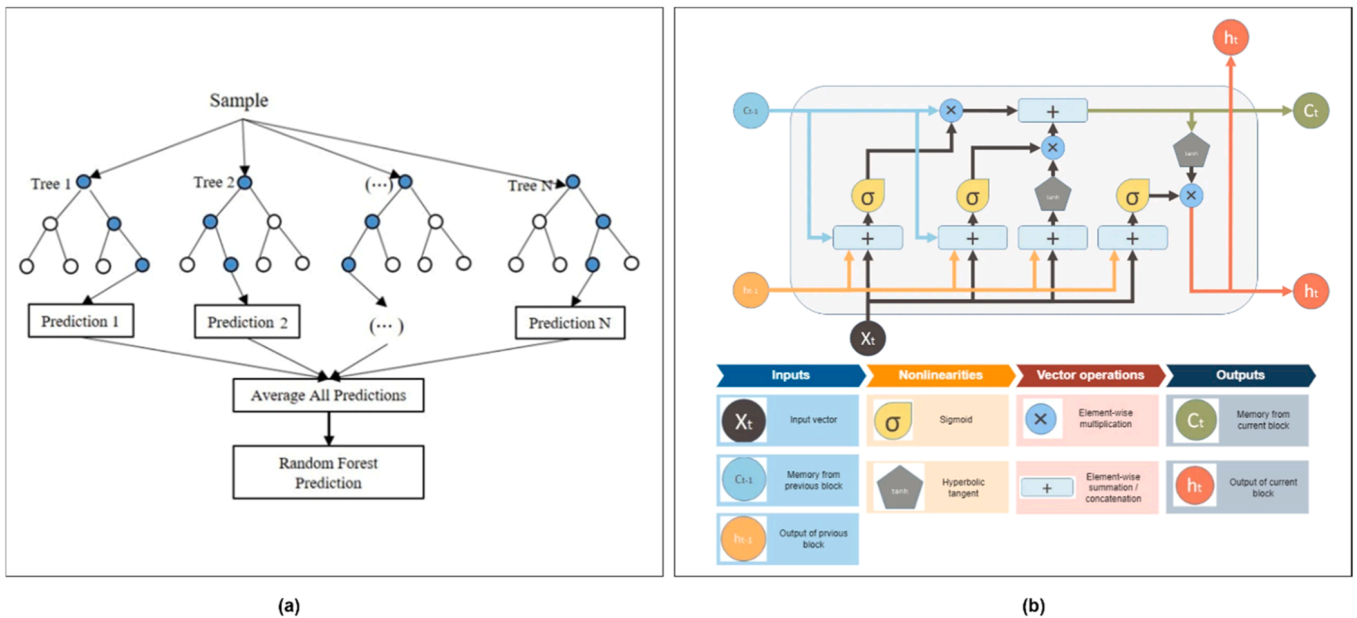

**Fig. 6.** (a) The general architecture of RF and (b) LSTM.

In this study, hyperparameter tuning was systematically performed for the RF, SVR, and LSTM models to ensure optimal predictive performance. For the RF model, a grid search approach was employed to explore key hyperparameters such as the number of trees (n_estimators), maximum tree depth (max_depth), and the minimum number of samples required to split a node (min_samples_split). The SVR model was fine-tuned by varying the regularisation parameter (C), the epsilon margin (epsilon), and kernel-specific parameters such as gamma for the radial basis function (RBF) kernel. For the LSTM model, hyperparameter optimisation involved adjusting the number of hidden units, number of layers, learning rate, batch size, and number of training epochs. All tuning procedures were conducted within a cross-validation framework to mitigate overfitting and ensure robust model selection. The best-performing hyperparameter combinations were selected based on cross-validated performance metrics and subsequently used for final model evaluation.

After verifying the linearity of the collected dataset, a "MinMax" scaling technique was applied to standardise the data, which was used for forecasting CWG. The primary focus of feature scaling was to address ambiguity and the presence of negative values in the dataset, utilising an appropriate scaling methodology. In this study, the Scikit-learn library in Python was employed to implement the "MinMax" scaler, which normalised the feature values between 0 and 1. Initially, the variables were reshaped into 2D arrays with dimensions (756, 1) to meet the scaler's input requirements. A range of 0 to 1 was specified, and each feature was individually adjusted through the fit_transform() function, which integrates the fitting and transformation steps. This function identified the range boundaries of the dataset, applying a proportional transformation to scale the data while preserving feature variability. The process ensured that the dataset was uniformly scaled, effectively mitigating issues such as inconsistencies or the presence of negative values. The following formula was employed to ensure that each feature value was scaled within the desired range.

$$z = (x - \min(x))/(\max(x) - \min(x)) \qquad (1)$$

Where $z$ represents the scaled value, and $\min(x)$ and $\max(x)$ denote the minimum and maximum values of attribute $x$ in the dataset, respectively.

### 3.5. Evaluation metrics

Evaluation metrics serve to determine the variation between observed and predicted values, providing insights into the model's effectiveness. This study applied four evaluation matrices, including the Coefficient of Determination ($R^2$), Root Mean Square Error (RMSE), Mean Absolute Error (MAE), Mean Absolute Percentage Error (MAPE), and accuracy ( %) to assess the predictive accuracy and reliability of RF, SVR, and LSTM models, highlighting each model's effectiveness and robustness in forecasting cattle weight trends.

The $R^2$ quantifies how well predicted values capture the variance in observed data, ranging from $-\infty$ to 1. An $R^2$ close to 1 indicates a strong model fit, while values near zero or negative reflect poor predictive performance [49,50,43]. Formulas (2) to (4) detail the mathematical expressions applied in assessing the $R^2$ evaluation metric.

$$R^2 = 1 - \left(\frac{RSS}{TSS}\right) \qquad (2)$$

$$RSS = \sum_{i=1}^{n} (y_i - \widehat{y_i})^2 \qquad (3)$$

$$TSS = \sum_{i=1}^{n} (y_i - \overline{y})^2 \qquad (4)$$

Where TSS refers to the total sum of squares, RSS denotes the residual sum of squares, $y_i$ indicates the observed value, $\widehat{y_i}$ represents the estimated value and $\overline{y}$ signifies the mean of the actual values.

RMSE quantifies prediction error by first computing the squared differences between predicted and actual values, then calculating their mean, followed by taking the square root. It emphasises larger errors, making it suitable for detecting substantial deviations between predicted and actual values [43]. The following formula is used to compute RMSE.

$$RMSE = \sqrt{\frac{\sum_{n=1}^{N}(\widehat{r_n} - r_n)^2}{N}} \qquad (5)$$

Where $\widehat{r_n}$ denotes the estimated rating, $r_n$ indicates the observed rating in the test dataset, and $N$ signifies the sample size of the testing dataset.

MAE is utilised to find out the sum of the absolute value of the prediction error. Initially, it computes the total by summing the absolute differences between the actual and predicted values, thus disregarding any negative signs. Subsequently, the mean is determined by averaging these computed absolute values. Consequently, the MAE values exhibit a linear variation because these are derived from the absolute differences between actual and predicted outcomes ([43]; W [51]). The formula below is implemented for MAE.

$$MAE = \frac{1}{n} \sum_{i=1}^{n} |y_i - x_i| \qquad (6)$$

Where $n$ denotes the number of test dataset samples, $y_i$ stands for the predicted outcome and $x_i$ represents the actual observed value.

MAPE expresses prediction accuracy as a percentage, offering a scale-independent interpretation of model performance. It provides an intuitive understanding of average prediction errors relative to the actual values, making it a preferred metric in many real-world forecasting tasks where interpretability and scale neutrality are crucial. The formula below is implemented for the accuracy metric.

$$MAPE = \frac{1}{n} \sum_{i=1}^{n} \left[\frac{y_i - \widehat{y_i}}{y_i}\right] \times 100 \qquad (7)$$

where $y_i$ stands for the observed values, $\widehat{y_i}$ corresponds to the estimated outcomes, and $n$ represents the sample size used in the evaluation.

In regression analysis, accuracy ( %) serves as a benchmark for assessing the precision of model predictions against actual outcomes. This metric is calculated as the proportion of accurate predictions, normalised as a percentage, with values closer to 100 % signifying superior predictive performance. Accuracy offers an interpretable measure for understanding how well the model generalises to new data and identifies key trends to evaluate regression models in practical applications. The formula below is implemented for the accuracy metric.

$$Accuracy(\%) = \left(1 - \frac{\sum_{i=1}^{n}|y_i - \widehat{y_i}|}{\sum_{i=1}^{n}|y_i|}\right) \times 100 \qquad (8)$$

where $y_i$ represents the actual observed values, $\widehat{y_i}$ refers to the estimated outcomes, and $n$ reflects the number of data points in the test set.

## 4. Experimental results

The main purpose of this experiment was to compare the prediction performances of RF, LSTM and SVR models on MB-CWG forecasting. The performance evaluation of the models was conducted on four datasets with varying combinations of cattle's background information, age and weather factors. In this case, a 10-fold cross-validation method was employed, whereby each model was iteratively trained using 90 % of the data and tested on the remaining 10 %. This method enabled a robust assessment of model generalisability across multiple data partitions. Tables 4, 6, and 8 present the performance results of the training datasets, while Tables 5, 7, and 9 show the outcomes of the testing datasets. The analysis considered several key metrics, including $R^2$, RMSE, MAE, MAPE and accuracy (%).

As shown in Tables 4 and 5, the RF model demonstrated superior predictive performance across all dataset configurations. When both weather and age attributes were included, the RF model reached the highest training accuracy, with an $R^2$ value of 0.973, RMSE of 0.040, and an accuracy of 96.75 %. The testing results closely aligned with the training outcomes, achieving an $R^2$ of 0.970 and 96.70 % accuracy. Even in scenarios where either weather or age data were excluded, the model retained predictive ability, with testing $R^2$ values remaining above 0.939 and accuracy above 93.70 %. Notably, even the most limited dataset, which lacked both age and weather factors, still resulted in respectable performance, showing the model's resilience to reduced feature input. These findings suggest that RF consistently maintains high accuracy and stability under varying data conditions.

The performance of the LSTM model is summarised in Tables 6 and 7. The performance of the LSTM model also reflected high predictive capacity, particularly when both weather and age variables were present. Under this full-feature configuration, the model produced a training $R^2$ of 0.969 and an accuracy of 95.04 %, with corresponding testing results of 0.966 for $R^2$ and the same accuracy percentage. When only weather or age factors were used, performance slightly declined but remained strong, with testing $R^2$ values of 0.949 and 0.945, respectively, and accuracy rates above 93 %. However, when both age and weather information were omitted, the model's accuracy dropped to 92.53 %, with an $R^2$ of 0.932.

As presented in Tables 8 and 9, the SVR model demonstrated moderate predictive capability relative to the RF and LSTM models. The strongest performance was observed when both age and weather variables were included in the dataset, resulting in a training $R^2$ of 0.963 and an accuracy of 94.90 %. In comparison, the corresponding testing results showed an $R^2$ of 0.959 and an accuracy of 94.89 %. The predictive performance declined when one or both feature categories were removed. Notably, in the absence of both age and weather information, the testing $R^2$ dropped to 0.908, with accuracy falling to 91.79 %.

This experiment compares the RMSE values of RF, LSTM, and SVR models for MB-CWG forecasting under different training dataset configurations. Fig. 7 shows the RMSE values for all models. RF had the lowest RMSE across all datasets, indicating its superior predictive accuracy. LSTM performed competitively but showed slightly higher RMSE than RF. SVR exhibited the highest RMSE, particularly for

**Table 4**
Accuracy assessment of the RF model on the training dataset.

| Evaluation Metrics | Dataset Including Weather and Age Factors | Dataset With Weather Factors (Excluding Age Factor) | Dataset With Age Factor (Excluding Weather Factors) | Dataset Excluding Age and Weather Factors |
|---|---|---|---|---|
| $R^2$ | 0.973 | 0.963 | 0.957 | 0.944 |
| RMSE | 0.040 | 0.046 | 0.050 | 0.055 |
| MAE | 0.033 | 0.037 | 0.044 | 0.052 |
| MAPE | 0.034 | 0.038 | 0.044 | 0.053 |
| Accuracy % | 96.75 % | 94.72 % | 94.47 % | 93.78 % |

**Table 5**
Accuracy assessment of the RF model on the testing dataset.

| Evaluation Metrics | Dataset Including Weather and Age Factors | Dataset With Weather Factors (Excluding Age Factor) | Dataset With Age Factor (Excluding Weather Factors) | Dataset Excluding Age and Weather Factors |
|---|---|---|---|---|
| $R^2$ | 0.970 | 0.963 | 0.951 | 0.939 |
| RMSE | 0.043 | 0.050 | 0.062 | 0.060 |
| MAE | 0.036 | 0.041 | 0.054 | 0.056 |
| MAPE | 0.038 | 0.042 | 0.055 | 0.057 |
| Accuracy % | 96.70 % | 94.69 % | 94.38 % | 93.70 % |

**Table 6**
Accuracy assessment of the LSTM model on the training dataset.

| Evaluation Metrics | Dataset Including Weather and Age Factors | Dataset With Weather Factors (Excluding Age Factor) | Dataset With Age Factor (Excluding Weather Factors) | Dataset Excluding Age and Weather Factors |
|---|---|---|---|---|
| $R^2$ | 0.969 | 0.951 | 0.949 | 0.939 |
| RMSE | 0.043 | 0.053 | 0.057 | 0.062 |
| MAE | 0.035 | 0.045 | 0.058 | 0.067 |
| MAPE | 0.036 | 0.045 | 0.059 | 0.069 |
| Accuracy % | 95.04 % | 94.52 % | 93.35 % | 92.54 % |

**Table 7**
Accuracy assessment of the LSTM model on the testing dataset.

| Evaluation Metrics | Dataset Including Weather and Age Factors | Dataset With Weather Factors (Excluding Age Factor) | Dataset With Age Factor (Excluding Weather Factors) | Dataset Excluding Age and Weather Factors |
|---|---|---|---|---|
| $R^2$ | 0.966 | 0.949 | 0.945 | 0.932 |
| RMSE | 0.044 | 0.053 | 0.058 | 0.063 |
| MAE | 0.035 | 0.045 | 0.058 | 0.067 |
| MAPE | 0.036 | 0.046 | 0.060 | 0.070 |
| Accuracy % | 95.04 % | 94.51 % | 93.34 % | 92.53 % |

**Table 8**
Accuracy assessment of the SVR model on the training dataset.

| Evaluation Metrics | Dataset Including Weather and Age Factors | Dataset With Weather Factors (Excluding Age Factor) | Dataset With Age Factor (Excluding Weather Factors) | Dataset Excluding Age and Weather Factors |
|---|---|---|---|---|
| $R^2$ | 0.963 | 0.936 | 0.923 | 0.916 |
| RMSE | 0.048 | 0.056 | 0.060 | 0.073 |
| MAE | 0.037 | 0.047 | 0.063 | 0.069 |
| MAPE | 0.038 | 0.049 | 0.065 | 0.072 |
| Accuracy % | 94.90 % | 94.40 % | 92.20 % | 91.80 % |

**Table 9**
Accuracy assessment of the SVR model on the testing dataset.

| Evaluation Metrics | Dataset Including Weather and Age Factors | Dataset With Weather Factors (Excluding Age Factor) | Dataset With Age Factor (Excluding Weather Factors) | Dataset Excluding Age and Weather Factors |
|---|---|---|---|---|
| $R^2$ | 0.959 | 0.932 | 0.920 | 0.908 |
| RMSE | 0.048 | 0.056 | 0.060 | 0.074 |
| MAE | 0.037 | 0.047 | 0.064 | 0.071 |
| MAPE | 0.039 | 0.049 | 0.065 | 0.073 |
| Accuracy % | 94.89 % | 94.40 % | 92.20 % | 91.79 % |

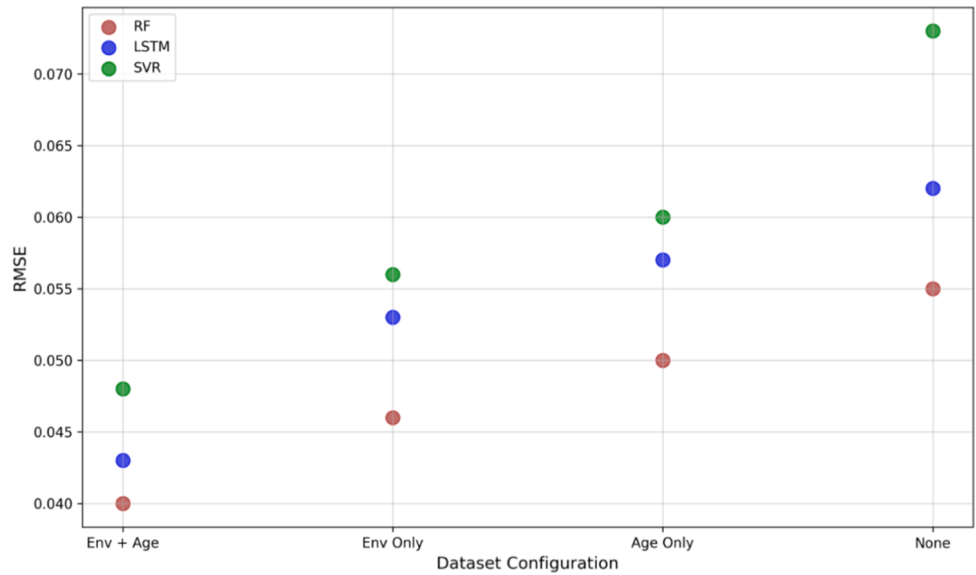

**Fig. 7.** RMSE values for all models.

datasets without weather or age factors.

The experiment also evaluated $R^2$ and MAE for RF, LSTM, and SVR models, as shown in Fig. 8(a) and Fig. 8(c). The heatmaps highlight how these metrics vary across different training datasets. RF consistently achieved the highest $R^2$ and the lowest MAE values, confirming its robustness. LSTM followed closely, while SVR demonstrated more variability in both metrics, particularly for datasets excluding critical features.

Fig. 9(a) presents a red dashed line representing perfect agreement, which acts as a basis for measuring the accuracy of the RF model. The close alignment of blue scatter points around this line emphasises the model's predictive effectiveness, with minor deviations suggesting occasional inconsistencies or outliers. Meanwhile, Fig. 9(b) demonstrates the alignment between actual and predicted cattle weights across months. Although the model tracks overall trends well, minor variations in May and August indicate challenges in forecasting, possibly influenced by external factors.

## 5. Discussion

The results of this study demonstrate the effectiveness of RF, LSTM, and SVR models in forecasting MB-CWG. Among the models evaluated, RF consistently delivered the highest accuracy and lowest error metrics across all training and testing datasets. Its performance suggests that RF is well-suited for handling complex datasets with diverse feature combinations. The inclusion of weather and age factors further enhanced RF's predictive capabilities, underscoring the importance of incorporating diverse data attributes in cattle weight forecasting. This outcome aligns with findings from earlier studies, such as Awasthi et al. [52], who also reported superior performance of tree-based ensemble models in cattle weight prediction. However, the current study extends that work by incorporating a broader feature space and evaluating models under various feature subsets, including age-only, weather-only, and combined inputs. While Ruchay et al. [53] evaluated ML models such as RF, Gradient Boosting, and SVM for Hereford cattle, the current study extends this work by testing time-based models like LSTM. It supports the effectiveness of RF and offers new insights by exploring prediction strategies suited to dynamic pasture environments.

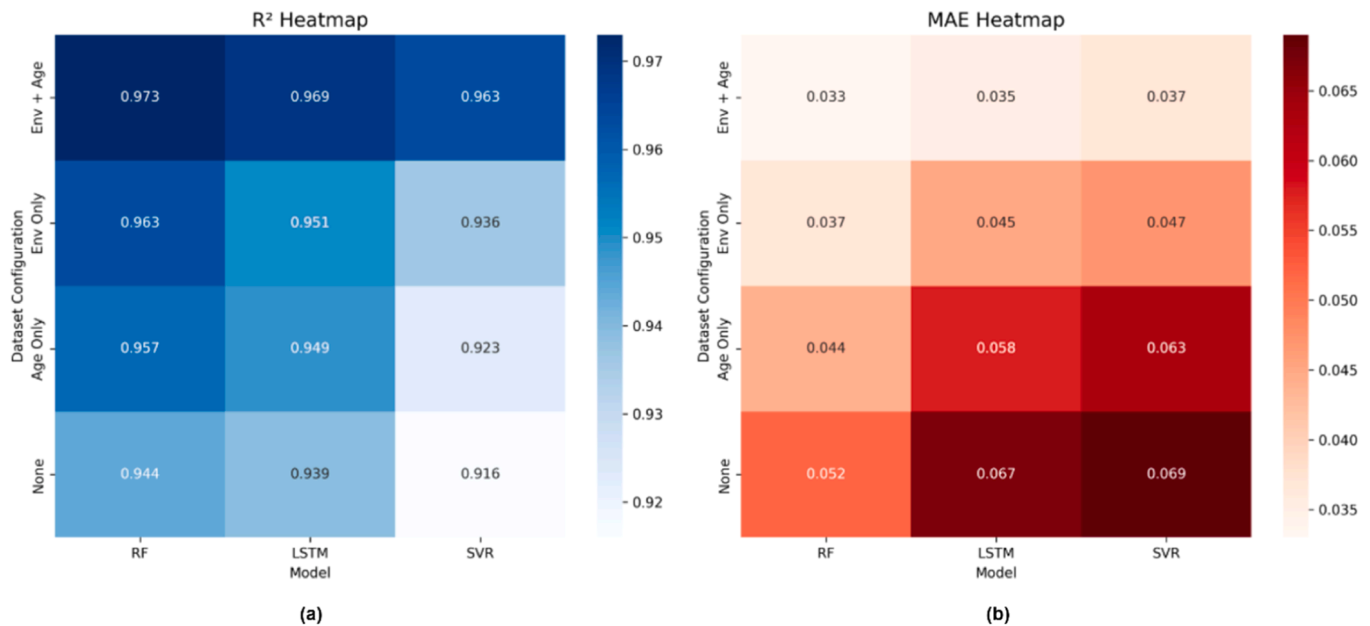

**Fig. 8.** (a) $R^2$ heatmap and (b) MAE heatmap.

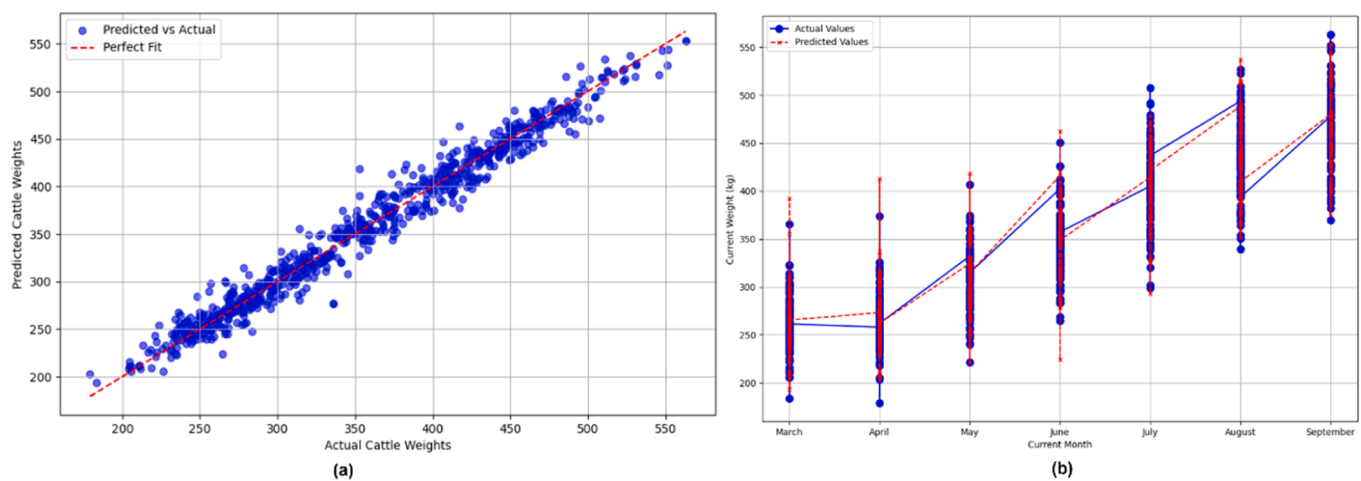

**Fig. 9.** (a) RF model's predictive accuracy with scatter plot, (b) comparison of monthly trends in cattle weights.

The LSTM model demonstrated strong performance when all input features were present, leveraging its strength in capturing temporal dependencies. However, a notable decline in accuracy and an increase in RMSE were observed when key predictors were removed, indicating sensitivity to feature dimensionality. This sensitivity suggests that while LSTM may be powerful for time-series applications, it requires complete and high-quality input to perform effectively. The study by Ruchay, Kober, Dorofeev, Kolpakov, Gladkov, et al. (2022) did not explore deep learning models such as LSTM, which positions the current research as an important contribution by addressing both traditional and advanced neural approaches for cattle weight prediction.

SVR produced acceptable results under simpler conditions but failed to maintain accuracy when faced with reduced or complex feature sets. This limitation is consistent with its known sensitivity to non-linear patterns. Such performance constraints have also been noted in prior research, reinforcing the idea that SVR may not be the most suitable choice for datasets with high variability or complex interactions. While Ruchay, Kober, Dorofeev, Kolpakov, Gladkov, et al. (2022) included SVM in their comparative analysis, they reported lower predictive power for SVM compared to RF and Boosted Trees, which supports the outcomes of this study.

The findings highlight the importance of model selection based on dataset characteristics. RF emerges as the most accurate option, while LSTM's strength lies in time-based analysis. SVR, though less adaptable, may serve as an efficient baseline model for less complex forecasting tasks. Furthermore, the influence of weather variables and cattle age on growth rates was pronounced, highlighting the need to incorporate these factors into predictive models.

Environmental and physiological variables also played a significant role in forecasting cattle weight gain. The analysis showed that cattle raised under moderate temperature and humidity conditions exhibited more consistent and higher weight gains than those exposed to extreme environmental stressors. This result suggests the potential for interventions like shade provision or humidity regulation to improve growth. The findings also demonstrated that younger cattle respond more positively to optimal conditions, suggesting the need for growth-phase-specific strategies, such as adjusted feeding programs or tailored housing designs.

## 6. Conclusions

This research compared the performance of RF, SVR, and LSTM models to predict the one-month advanced weight gain of mob-based cattle. RF proved to be the most accurate and reliable model due to its ability to handle a variety of features, making it ideal for use in commercial cattle management. LSTM performed well with data involving time patterns, showing its potential for time-series applications. SVR worked well for simple tasks but struggled with more complex data.

Accurate prediction of MB-CWG depends on the integration of key biological and environmental variables, such as age and weather data. The current study confirms that excluding these variables reduces model effectiveness, highlighting their critical role in capturing cattle growth trends. In addition, an automated pre-processing tool was also developed during the study, offering a novel approach for creating a benchmark dataset tailored to MB-CWG prediction models. Predictive modelling is recommended as a practical solution for assisting farmers in managing resources efficiently while supporting reliable forecasting of one-month advanced weight gain in mob-based cattle systems.

Despite the promising results, the study encountered several limitations. The dataset was restricted to a specific context, with samples sourced from a single location. This geographic and environmental homogeneity may reduce the generalisability of the findings. Additionally, the investigation was confined to a few predictive variables, leaving unexplored several biological, nutritional, and management-related factors that could further influence cattle growth. Despite these constraints, the research demonstrates the potential of ML for forecasting cattle weight and highlights the value of continued innovation in predictive modelling. Such tools have the potential to transform how producers manage livestock growth, allocate resources, and improve productivity in a data-driven agricultural future.

Future work should aim to validate the model across multiple farms operating under diverse environmental and management conditions. Expanding the feature set to include feed intake, pasture quality, genetic factors, and health records could enhance model interpretability and prediction power. A shift from monthly to daily prediction intervals may also provide producers with more granular insights, supporting real-time decision-making in livestock management. Furthermore, integrating the complementary strengths of RF and LSTM into a hybrid architecture could offer a more balanced and effective predictive system capable of handling both static and sequential patterns in complex datasets.

**Funding sources**

This project was supported by funding from Food Agility CRC Ltd under the Commonwealth Government CRC Program. The CRC Program supports industry-led collaborations between industry, researchers, and the community.

## CRediT authorship contribution statement

**Muhammad Riaz Hasib Hossain:** Writing – original draft, Visualization, Validation, Software, Resources, Methodology, Investigation, Formal analysis, Data curation, Conceptualization. **Md Rafiqul Islam:** Writing – review & editing, Supervision, Project administration, Funding acquisition. **S.R. McGrath:** Writing – review & editing, Supervision, Funding acquisition. **Md Zahidul Islam:** Writing – review & editing, Supervision. **David Lamb:** Writing – review & editing, Supervision, Funding acquisition.

## Declaration of competing interest

The authors declare that they have no known competing financial interests or personal relationships that could have influenced the work reported in this paper.

## Data availability

Data will be made available on request.